\pdfoutput=1
\documentclass[letterpaper, 10 pt, conference]{ieeeconf}  
\usepackage{amsmath}
\usepackage{float}

\usepackage{subcaption}
\usepackage{graphicx}
\usepackage{amssymb} 
\IEEEoverridecommandlockouts                              

\title{\LARGE \bf
GPS Denied IBVS-Based Navigation and Collision Avoidance of UAV Using a Low-Cost RGB Camera*
}

\author{Xiaoyu Wang$^{1}$, Yan Rui Tan$^{2}$, William Leong$^{2}$, Sunan Huang$^{2}$, Rodney Teo$^{2}$  and Cheng Xiang$^{1}$ 
\thanks{*This work was not supported by any organization}
\thanks{$^{1}$Xiaoxu and Cheng are with Electrical and Computer Engineering Dept., National  University of Singapore, 4 Engineering Drive 3, Singapore 117583
        {\tt\small e1350382@u.nus.edu,elexc@nus.edu.sg}}%
\thanks{$^{2}$Yan Rui, William, Sunan and Rodney are with the National University of Singapore, Singapore
        {\tt\small \{yanrui, william.leong, tslhs, tsltshr\}@nus.edu.sg}}%
}

\begin{document}

\maketitle
\thispagestyle{empty}
\pagestyle{empty}

\begin{abstract}

This paper proposes an image-based visual servoing (IBVS) framework for UAV navigation and collision avoidance using only an RGB camera. While UAV navigation has been extensively studied, it remains challenging to apply IBVS in missions involving multiple visual targets and collision avoidance. The proposed method achieves navigation without explicit path planning, and collision avoidance is realized through AI-based monocular depth estimation from RGB images. Unlike approaches that rely on stereo cameras or external workstations, our framework runs fully onboard a Jetson platform, ensuring a self-contained and deployable system. Experimental results validate that the UAV can navigate across multiple AprilTags and avoid obstacles effectively in GPS-denied environments.  
\end{abstract}

\section{INTRODUCTION}
Most UAV applications depend on position estimation provided by global positioning systems (GPS). However, GPS is often unavailable in indoor, mountainous, or forest environments, motivating the use of computer vision for UAV navigation. This paper focuses on image-based visual servoing (IBVS) with an onboard RGB camera.

Traditionally, visual servoing algorithms for UAVs are divided into
two classes: position-based visual servoing (PBVS)  and image-based visual servoing (IBVS) \cite{ fc_sh_2007, sh_pic_1996,sc_dhs2019,aao_ea_jds_prg_fc2022}. PBVS is used to reconstruct a global 3D pose of
the target using RGB, stereo vision, or depth information, etc. This introduces a limitation--global localization. IBVS uses the error between
detected and desired visual features to control motion. It is effective for missions where the visual features are within the camera's field of vision. 

In GPS-denied settings, in some cases, global localization is difficult to achieve by using a Visual-Inertial Odometry (VIO) 
or Simultaneous Localization and Mapping (SLAM). This often becomes a critical point of failure.  IBVS can be used for handling such kinds of environments, leading to a more robust result.  However, most existing IBVSs are for one target (stationary or moving target)  tracking (\cite{hja_my_jy_2016,hja_jy_2016,amm_hc_jlg_hw_2022,jl_hx_khl_jy_bl_2021,ly_xw_yz_zl_ls_2024}) or landing \cite{akk_ntc_mf_2024}. Navigation missions can involve multiple waypoints and require the UAV to switch among multiple waypoints to guide the UAV to achieve the navigation task.  Another challenge in the navigation problem is the handling of collision avoidance \cite{sh_rsht_kkt_2019}.  Most existing  IBVS results for navigation with collision avoidance use stereo cameras (or other sensors) to obtain depth information. For example, in \cite{ahah_akn_cvj_2015, hz_jz_xy_yl_lc_kz_sl_xh_2023}, the authors find a feasible trajectory in 3D Cartesian space which uses PBVS-like method for path planning;   in \cite{cdw_fpv_ns_gched_2022,cq_msjm_jc_hhtl_2024}, the authors use stereo cameras to obtain the depth and do the navigation; in \cite{ly_xw_yz_zl_ls_2024,dl_hl_hjk_2011}, the authors use the UAV position, LOS information or exact depth for collision avoidance. Although \cite{cdw_fpv_ns_gched_2022,cq_msjm_jc_hhtl_2024} present the navigation method based on the stereo cameras, collision avoidance is not discussed. These results need to have the exact UAV position or relative position information. It is not realistic to use exact 3D position information to handle collision avoidance, for the depth information is not always accurate for stereo cameras, especially in a moving drone with stereo cameras, depending on several factors like calibration, lighting, camera baseline, and distance, etc.  It is preferable to use a low-cost RGB camera for sensing applications for navigation with collision avoidance.  The authors in \cite{pw_sh_wll_zm_ss_rsht_2022} utilized a low-cost RGB camera for collision avoidance. However, it requires prior knowledge of the obstacle’s size. In addition, its navigation uses GPS information.

In this paper, we present an IBVS-based navigation method, where the collision avoidance is achieved partially by using  AI estimation based on an RGB camera only. The proposed method can work in a denied environment without a map, VIO, SLAM, LiDAR,   or stereo camera. 
Unlike approaches that offload heavy perception to external high-performance workstations, our entire framework (including MiDaS inference) runs fully onboard a Jetson platform, ensuring a self-contained and deployable system. 
The main contributions include
\begin{itemize}
\item A multiple visual target set is introduced for guiding the UAV for navigation. For each target, IBVS is used for achieving the tracking control. A switching law is developed to complete the transition from one target to the other. During the navigation, only RGB is used without a path planning requirement.
\item The obstacles are detected by the depth map estimated by AI learning using RGB images, without requiring prior knowledge or explicit encoding as in \cite{rmb_cz_qn_2023}. 
In contrast, the navigation gates are encoded with visual AprilTags, which are identified by IBVS.  

\item A collision avoidance law is developed by using the depth percentage thresholds.  During the collision avoidance, the accurate position of UAV \cite{dl_hl_hjk_2011} or the relative position information of the obstacles \cite{ahah_akn_cvj_2015, hz_jz_xy_yl_lc_kz_sl_xh_2023} is not required. 
\end{itemize}

The paper is organized as follows. Section II describes the detailed methodology designed. Section III presents the test results using the proposed method and gives a comparison with existing methods. Finally, the conclusion is given in Section IV.

\section{METHODOLOGY}
\begin{figure}[t]
    \centering
    \includegraphics[width=0.5\linewidth]{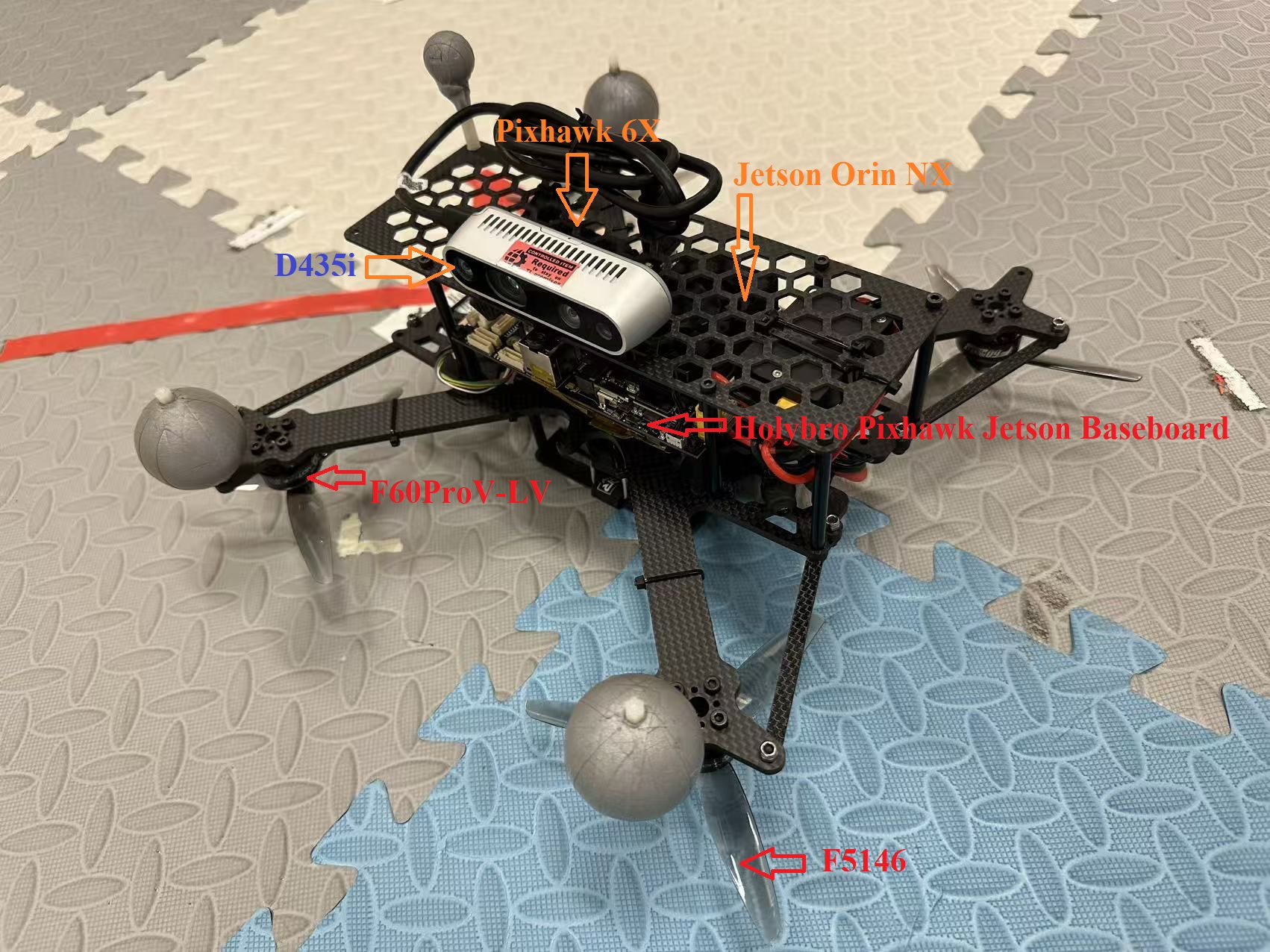}
    \caption{Quadrotor drone}
    \label{fig:drone}
\end{figure}

The designed drone, as shown in Fig.~\ref{fig:drone}, weighs 1.1485 kg with a thrust-to-weight ratio of 4:1, where the arm length is 15.5cm, four T-motor F60 ProV-lv are used,  the propellers are F5146, and the battery is a 6-cell LiPo. The motors used are 2020KV. An Intel RealSense D435i camera is mounted on the drone and connected to the Jetson Orin baseboard.

\begin{figure*}[t]
    \centering
    \includegraphics[width=0.5\linewidth]{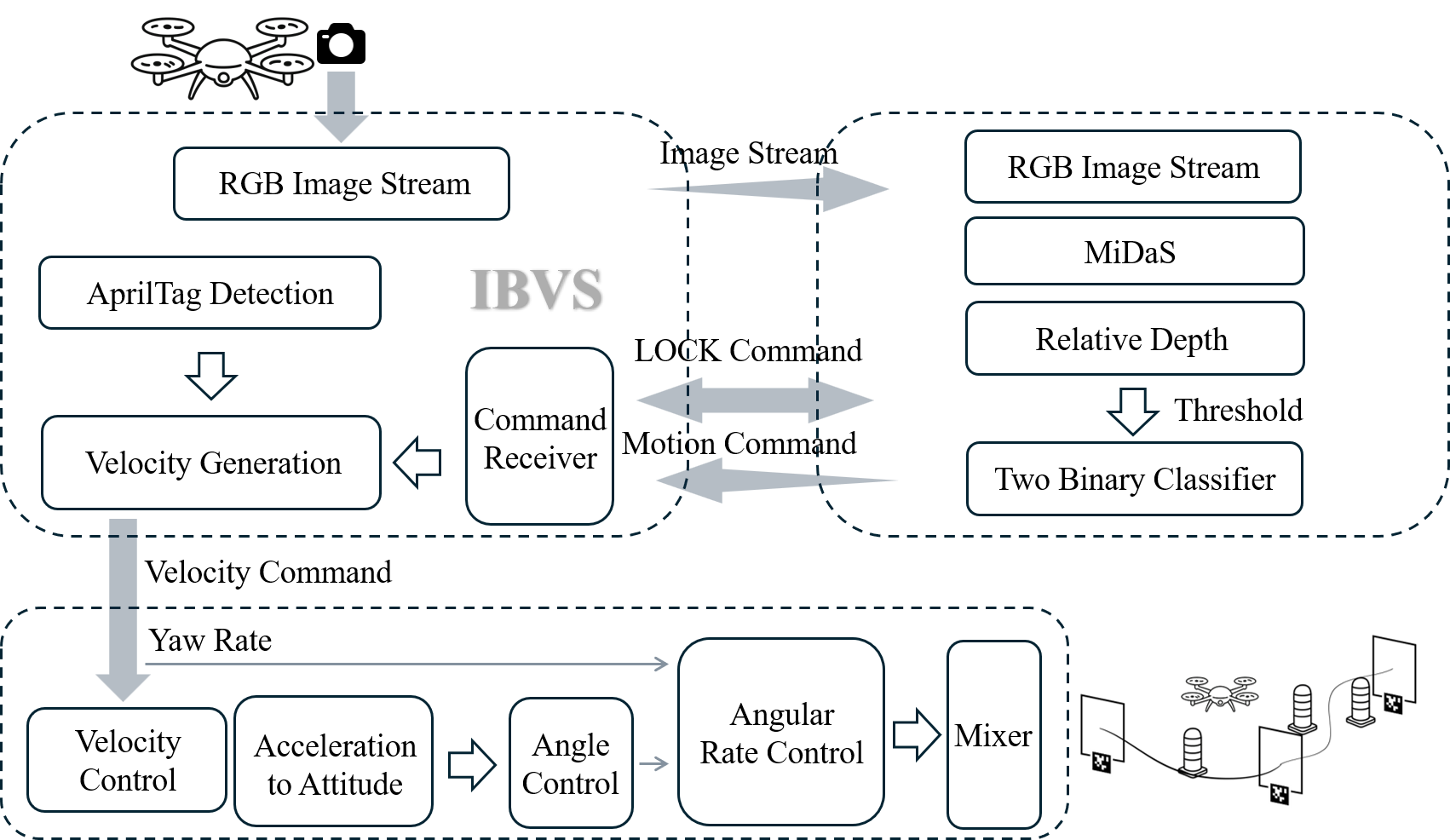}
    \caption{Overall technical pipeline of the proposed IBVS-based navigation and collision avoidance framework. 
    The left branch shows AprilTag detection and IBVS velocity control; the right branch shows MiDaS-based depth estimation and obstacle avoidance; the bottom illustrates PX4 flight control loops}
    \label{fig:method}
\end{figure*}

The proposed framework integrates image-based visual servoing (IBVS) with AI-driven depth estimation to achieve navigation and collision avoidance in GPS-denied environments using only a low-cost RGB camera. The overall architecture is shown in Fig.~\ref{fig:method} and consists of three main components: navigation via AprilTag tracking, collision avoidance using monocular depth estimation, and PX4-based flight control.

Navigation is designed around a sequence of visual landmarks placed along the desired path. AprilTags serve as the visual features, each associated with a gate that the UAV must pass through. IBVS is employed to track these tags using image moments, which provide robust visual features for translational and rotational motion control.

To enhance safety, a monocular depth estimation module based on the \texttt{openvino\_midas\_v21\_small} model that runs in parallel to IBVS. The incoming RGB stream is processed to generate a relative depth map, which is normalized and thresholded to produce a binary obstacle mask. By analyzing the mask's centroid and pixel distribution, the system determines the relative position of obstacles to the drone's field of view and generates appropriate obstacle avoidance commands.

The flight control module is implemented in PX4 and is divided into three cascaded loops. This layered structure ensures stable flight performance while also integrating IBVS velocity commands and yaw feedforward inputs for responsive maneuvers.

\subsection{IBVS-based navigation}
A standard navigation for a drone is to start from one point to the destination. In our project, the navigation is composed of three parts: a series of targets, target tracking, and a switching law:
\begin{itemize}
\item \underline{A set of targets}. According to the navigation mission, we use a set of multiple targets to construct a series of visual feature sequences from the starting point to the destination. Here, AprilTags are used as visual features. Each AprilTag representing an ID, $id_k (k=0,1,...,n)$, must be placed within camera detection range, where we use camera D435i, which has an optimal detection range of 4m. Thus, the distance between two targets should be less than 4m. Therefore, we have the following target points, 
\begin{eqnarray*}
   id_0(starting \;point), id_1, id_2,..., id_n (destination). 
\end{eqnarray*}
Each target corresponds to a gate. The next step is to track the target and pass through the gate.
\item \underline{Target tracking}. As shown in Fig. \ref{fig:method}, IBVS uses target detection to identify the visual features of AprilTag and generates the velocity control command. The main purpose of  IBVS is for target tracking. We use the image moment concept in IBVS.  The image moments $m_{ij}$ of pixels of visual features  are defined as
\begin{eqnarray}
m_{ij} = \iint_{O} x^i y^j \, dx \, dy
\end{eqnarray}
The discrete form is given by
\begin{eqnarray}
m_{ij} = \sum_{k=1}^{n} x_k^i y_k^j
\end{eqnarray}
where $n$ is the image points in the target. The centered moments $\mu_{ij}$ are given by
\begin{eqnarray}
\mu_{ij} = \sum_{k=1}^{n} (x_k - x_g)^i (y_k - y_g)^j
\end{eqnarray}
where
\begin{eqnarray}
x_g = \frac{m_{10}}{a},\quad y_g = \frac{m_{01}}{a}.
\end{eqnarray}
Now, the image features for the translational motion control are given by
\begin{eqnarray}
x_n=a_n{x_g}\\
y_n=a_n{y_g}\\
a_n=z*\sqrt{\frac{a*}{a}}
\end{eqnarray}
where  $z*$ is the desired normal distance of the camera from the object, and $a*$ is the desired value of $a$ and $a=\mu_{20}+\mu_{02}$. For the orientation of the yaw rate, it uses the polar coordinates $(\rho, \theta)$ and $arctan(1/\rho)$ is used as our visual feature. Referring to  the reference \cite{ot_fc_2005}, the  velocity command  $v_c=[v_x,v_y,v_z,w_z]^T$ corresponding respectively to the 3 linear  velocities $[v_x,v_y,v_z]$, and the yaw rate $w_z$, is given by
\begin{eqnarray}
v_c=-\lambda \hat{L}^{-1}(q-q^*)
\end{eqnarray}
where  $\hat{L}$ is the interaction matrix,  $q=[x_n,y_n,a_n, arctan(1/\rho) ]^T, $ and $q^*$ is the desired image features.  

\item \underline{Switching law}.  When the UAV is approaching the target, it cannot detect the target (D435i has a minimum detection distance of 0.3m). In this case, the switching law is used to handle the transition from one target to another. It includes the passing gate and the searching laws. The passing gate law uses 'Up' and 'Forward' commands to control the drone passing through the gate. After passing the gate, the drone will search for the next target. The searching law uses the following rules:
\begin{itemize}
\item If the target is located  on the left side of the drone,  it turns to the left and  searches for the target;
\item
If the target is located on the right side of the drone, it turns to the right and searches for the target.
\item
Similarly, we can use up or down for searching targets;
\item
If we don’t know the target direction, the drone will rotate 360 degrees for searching. 

\end{itemize}
\end{itemize}

\subsection{Collision avoidance}

The collision avoidance module relies on AI-based depth estimation and motion cues. 
First, a MiDaS network running on the Python side processes the RGB image stream to generate pseudo-depth maps. 
These maps are then normalized and thresholded to produce a binary obstacle mask. 
Second, based on the distribution of white pixels in the mask and the centroid position, the system determines whether an obstacle lies on the left, right, or center of the image. 
A corresponding command (\texttt{LEFT}, \texttt{RIGHT}, or \texttt{CENTER}) is transmitted via UDP to the C++ IBVS controller. 

Upon receiving the command, the IBVS module applies a LOCK/UNLOCK mechanism to ensure that each avoidance maneuver is executed to completion before the next decision is considered. 
Finally, the UAV executes motion strategies according to the detected obstacle location: if an obstacle is detected on the left, the UAV performs a right turn and forward motion; if an obstacle is on the right, a left turn is commanded. 
When the obstacle area is insufficient or centered, the UAV maintains forward flight or hovering.

For collision avoidance, we adopt a monocular depth estimation approach based on the MiDaS framework \cite{Ranftl2020, Ranftl2021, Ranftl2019}. 
Our MiDaS module is optimized with OpenVINO and executed directly on the Jetson onboard computer. This enables real-time inference ($\approx$4 FPS) within the UAV’s onboard computational budget, making the overall framework more practical for field deployment.
In particular, we utilize the \texttt{openvino\_midas\_v21\_small} model, which provides efficient inference on edge devices while maintaining competitive accuracy. 
The network is trained using a mixture of multiple datasets with scale- and shift-invariant loss functions, enabling zero-shot generalization across environments. 
Given an input RGB frame $I \in \mathbb{R}^{H \times W \times 3}$, the model predicts an inverse depth map 
\begin{eqnarray}
d(x,y) = f_{\theta}(I(x,y)),
\end{eqnarray}
where $(x,y)$ denotes image coordinates and $\theta$ are network parameters. 
Since monocular depth estimation is inherently ambiguous up to scale and shift, the predictions are aligned to ground truth (or used in a relative sense) via a least-squares fit:
\begin{eqnarray}
(s,t) = \arg\min_{s,t} \sum_{i=1}^{M} \left( s d_i + t - d_i^* \right)^2,
\end{eqnarray}
where $d_i$ and $d_i^*$ are predicted and reference disparities, respectively. 
This alignment yields scale- and shift-invariant predictions suitable for downstream control.

In deployment, each predicted depth map is normalized and thresholded to produce a binary obstacle mask:
\begin{eqnarray}
M(x,y) = \mathbb{1}\!\left[d(x,y) > \tau\right],
\end{eqnarray}
where $\tau$ is the raw depth threshold (empirically chosen as $\tau=900$ in our setup). 
The centroid and pixel distribution of $M(x,y)$ are analyzed to infer obstacle position relative to the UAV’s field of view. 

The combination of MiDaS-based pseudo-depth and IBVS control laws enables real-time obstacle avoidance using only a low-cost RGB camera, without requiring LiDAR, stereo camera, or GPS.

\subsection{Flight controller}
The flight controller has three loops as shown in Figure~\ref{fig:method}. The linear velocity command from  IBVS is used as the reference for the velocity control loop. This loop generates an acceleration command, which is converted into a quaternion command and a thrust command. The attitude control loop is used to generate the angular rate command, where the yaw rate command generated from IBVS is used as a feedforward term.  For the angular rate control, a PID controller is designed to generate torque control. Here, we use PX4 software, which uses a control allocation (Mixer) to translate the thrust and torque commands into individual rotor commands.

\section{Test Results}   

The experiment is conducted in an indoor test site designed for UAV visual servoing and obstacle avoidance. 

\subsection{Test environment}
On the drone, the Jetson Orin executes the algorithms through MAVSDK, which is an interface to communicate using MAVLink with the PX4 flight controller.  In the ground control station, we have  QGroundControl (QGC), which provides the flight status. 

\begin{figure}[thpb]
    \centering
    \begin{subfigure}[b]{0.35\linewidth}
        \centering
        \includegraphics[width=\linewidth]{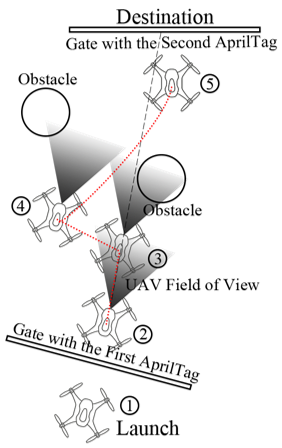}
        \caption{Top view diagram}
        \label{fig:top_view}
    \end{subfigure}
    \hspace{0.02\linewidth}
    \begin{subfigure}[b]{0.6\linewidth}
        \centering
        \includegraphics[width=\linewidth]{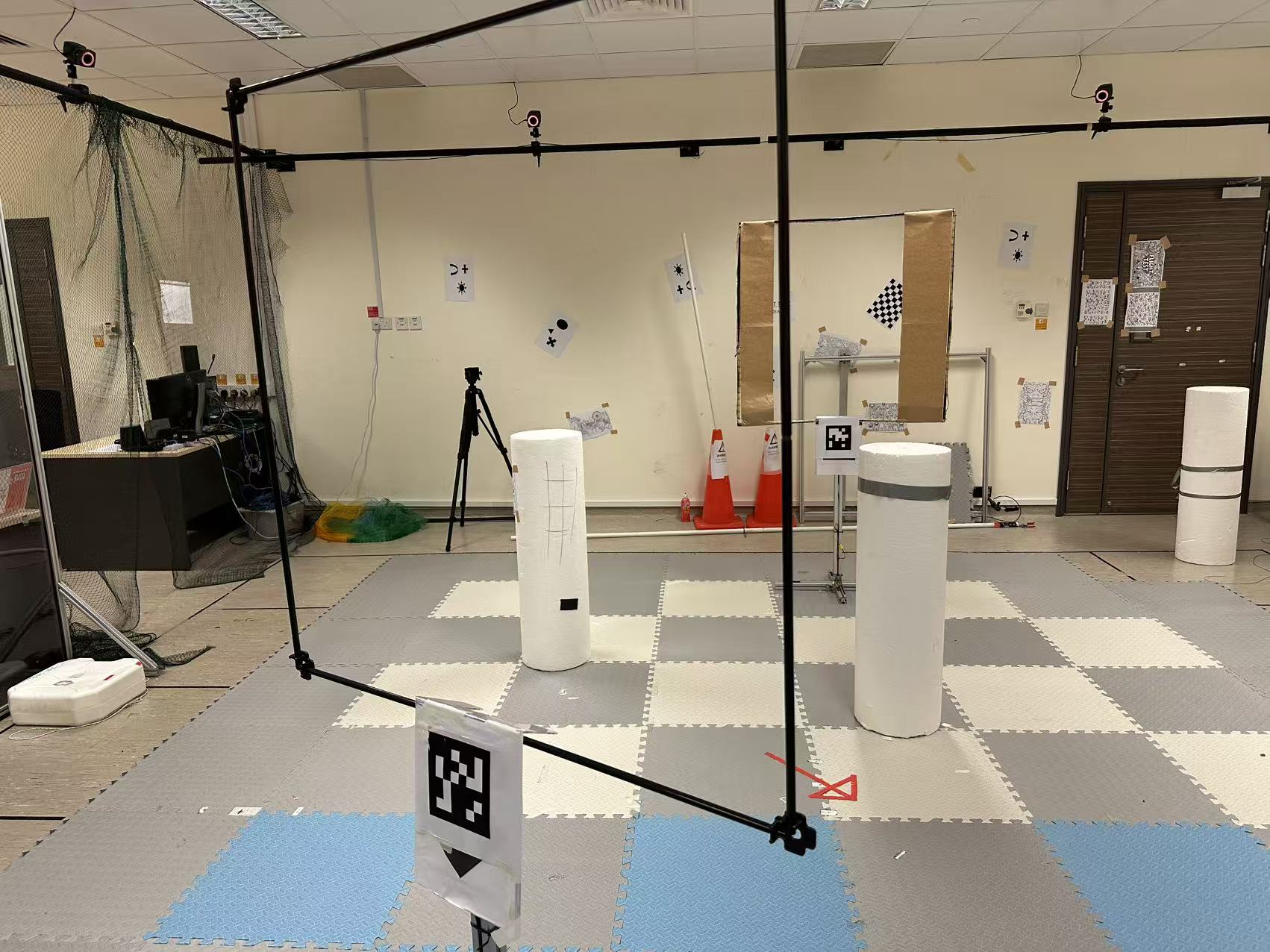}
        \caption{Actual experimental site}
        \label{fig:real_gate}
    \end{subfigure}
    \caption{Experimental site: Obstacles and gates associated with AprilTags: (a) top view diagram; (b) actual experimental site}
    \label{fig:gate}
\end{figure}

For the navigation mission, we use two AprilTags as visual targets to guide the drone: one guides the drone to an intermediate waypoint, and the other one to the destination. IBVS will be used to perform target tracking.  As shown in Fig.~\ref{fig:gate}, after takeoff, the UAV first aligns with and tracks the initial AprilTag. Once the image error falls below a predefined threshold, the UAV executes the gate-crossing maneuver. Upon passing through the gate, the UAV is required to search and identify the second AprilTag and continue tracking it while simultaneously avoiding two cylindrical obstacles positioned in the flight path. A VICON motion capture system is installed around the arena to provide ground truth measurements for evaluation.

The system is tested in this experiment. For the collision avoidance, the MiDaS-based obstacle detection and avoidance information are sent to the IBVS framework to enable the drone to avoid obstacles during the navigation process. With the MiDaS module, the drone successfully tracked consecutive AprilTags, passed through the designated gates, and avoided the cylindrical obstacles.

It is important to emphasize that all computations, including MiDaS inference, were performed onboard the Jetson platform. 
This contrasts with methods that rely on powerful external 
workstations, and validates that real-time UAV navigation and obstacle avoidance can be achieved within the onboard hardware constraints.

\begin{itemize}
\item \underline{Tracking the $1^{\mathrm{st}}$ AprilTag}: After taking off, the UAV successfully detected the first AprilTag and initiated image-based visual servoing (IBVS) for target tracking. As the servoing process proceeded, the tracking error gradually decreased, demonstrating the UAV’s ability to align with the visual target in real time. Once the error decreased below the designated threshold, the system determined that the target had been successfully tracked, which triggered the subsequent gate-crossing action.

\begin{figure*}[t]
    \centering
    \begin{subfigure}[b]{0.49\linewidth}
        \centering
        \includegraphics[width=\linewidth,height=4.2cm]{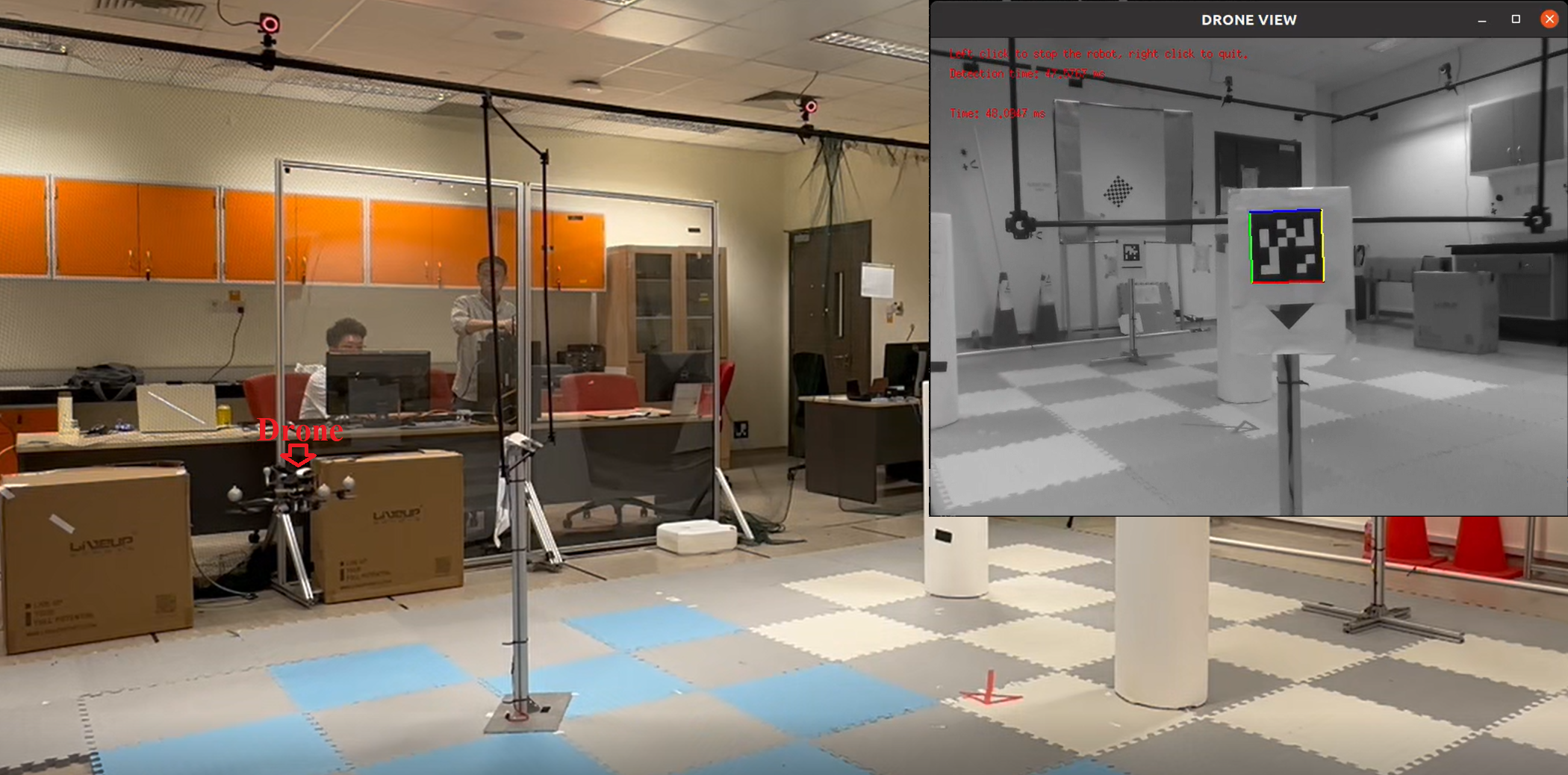}
        \caption{UAV tracks the first AprilTag}
        \label{fig:track1}
    \end{subfigure}
    \begin{subfigure}[b]{0.49\linewidth}
        \centering
        \includegraphics[width=\linewidth,height=4.2cm]{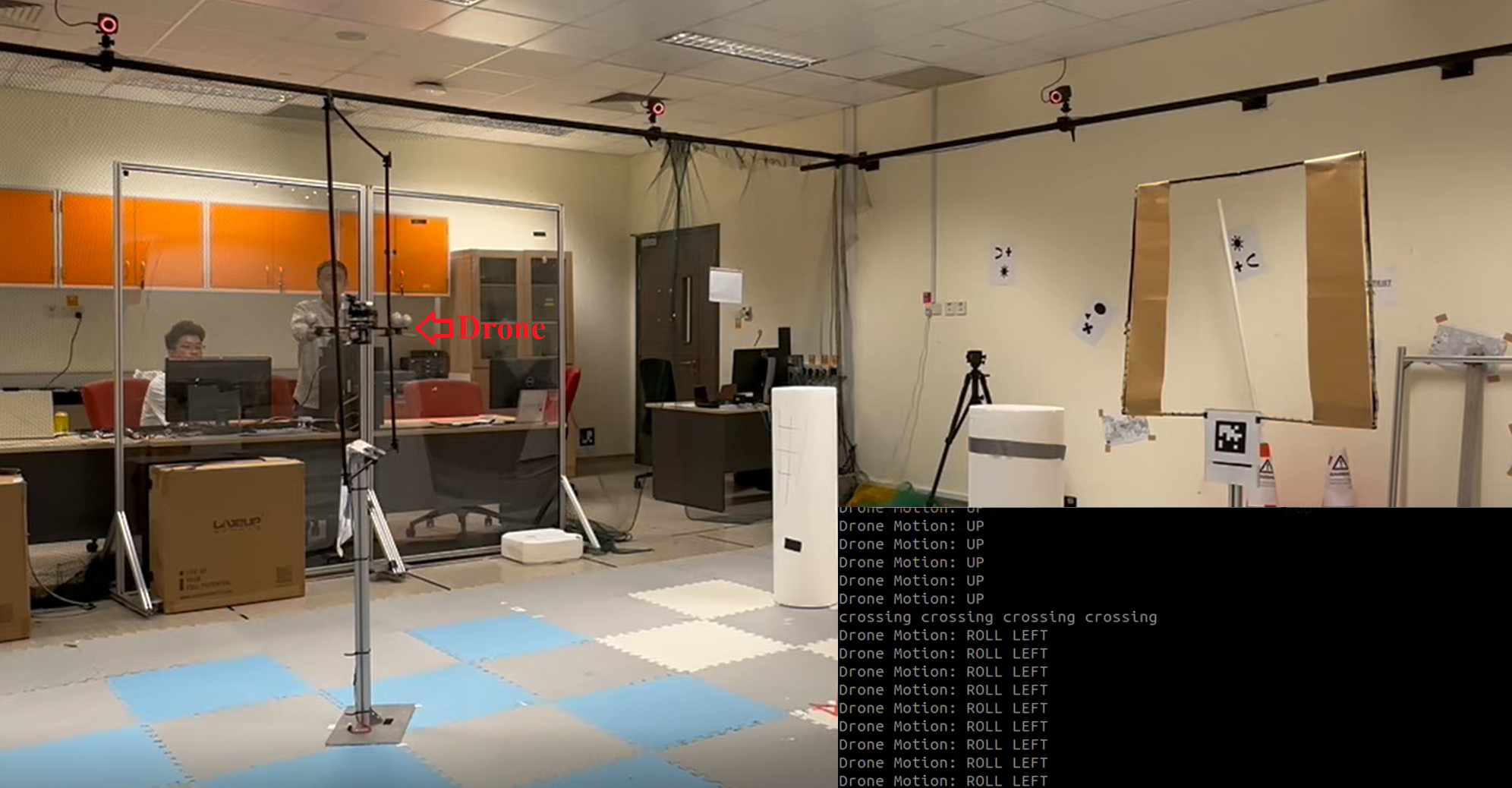}
        \caption{UAV crosses the gate}
        \label{fig:cross1}
    \end{subfigure}

    \begin{subfigure}[b]{0.49\linewidth}
        \centering
        \includegraphics[width=\linewidth,height=4.2cm]{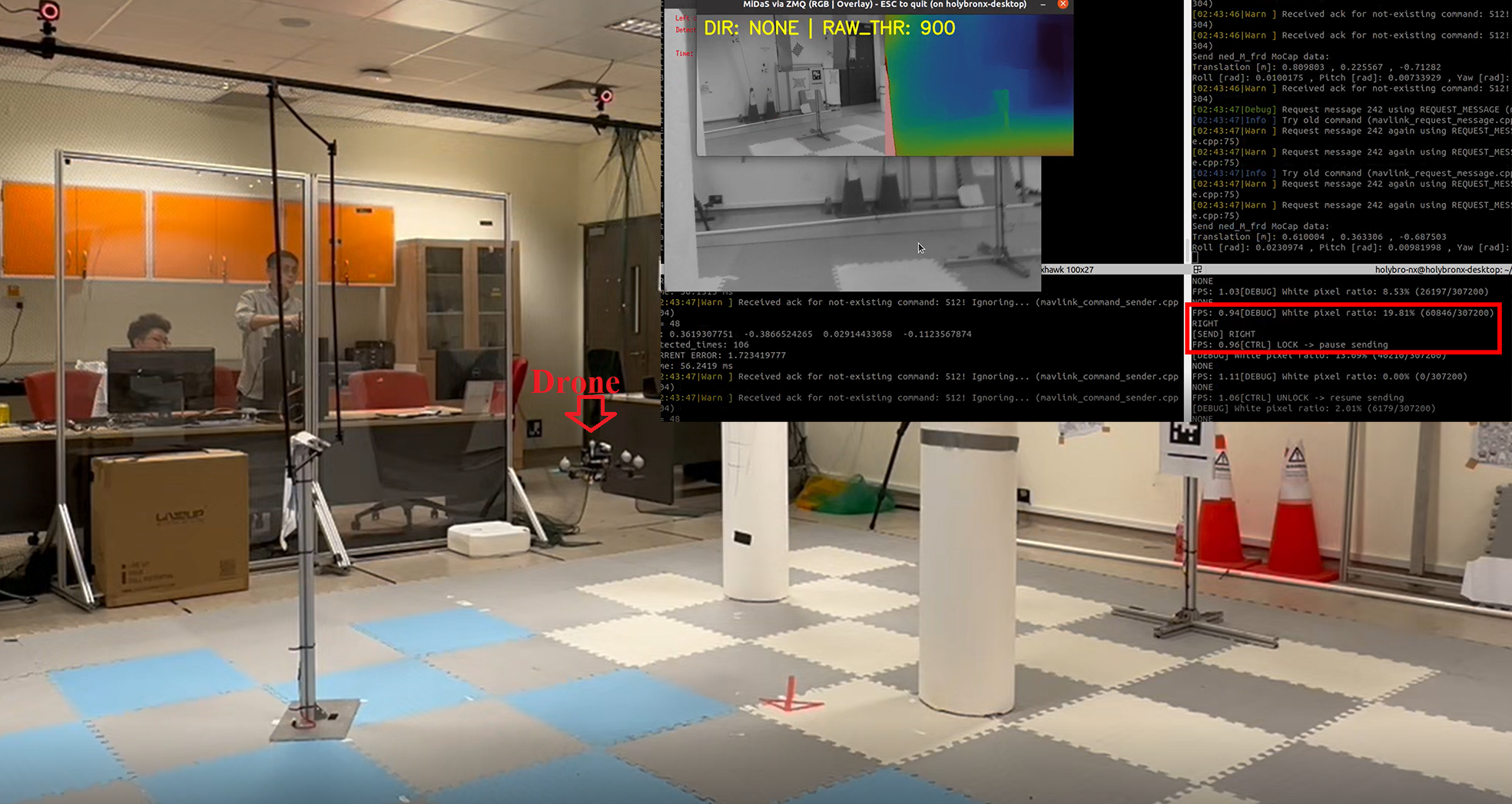}
        \caption{UAV tracks the second AprilTag}
        \label{fig:track2}
    \end{subfigure}
    \begin{subfigure}[b]{0.49\linewidth}
        \centering
        \includegraphics[width=\linewidth,height=4.2cm]{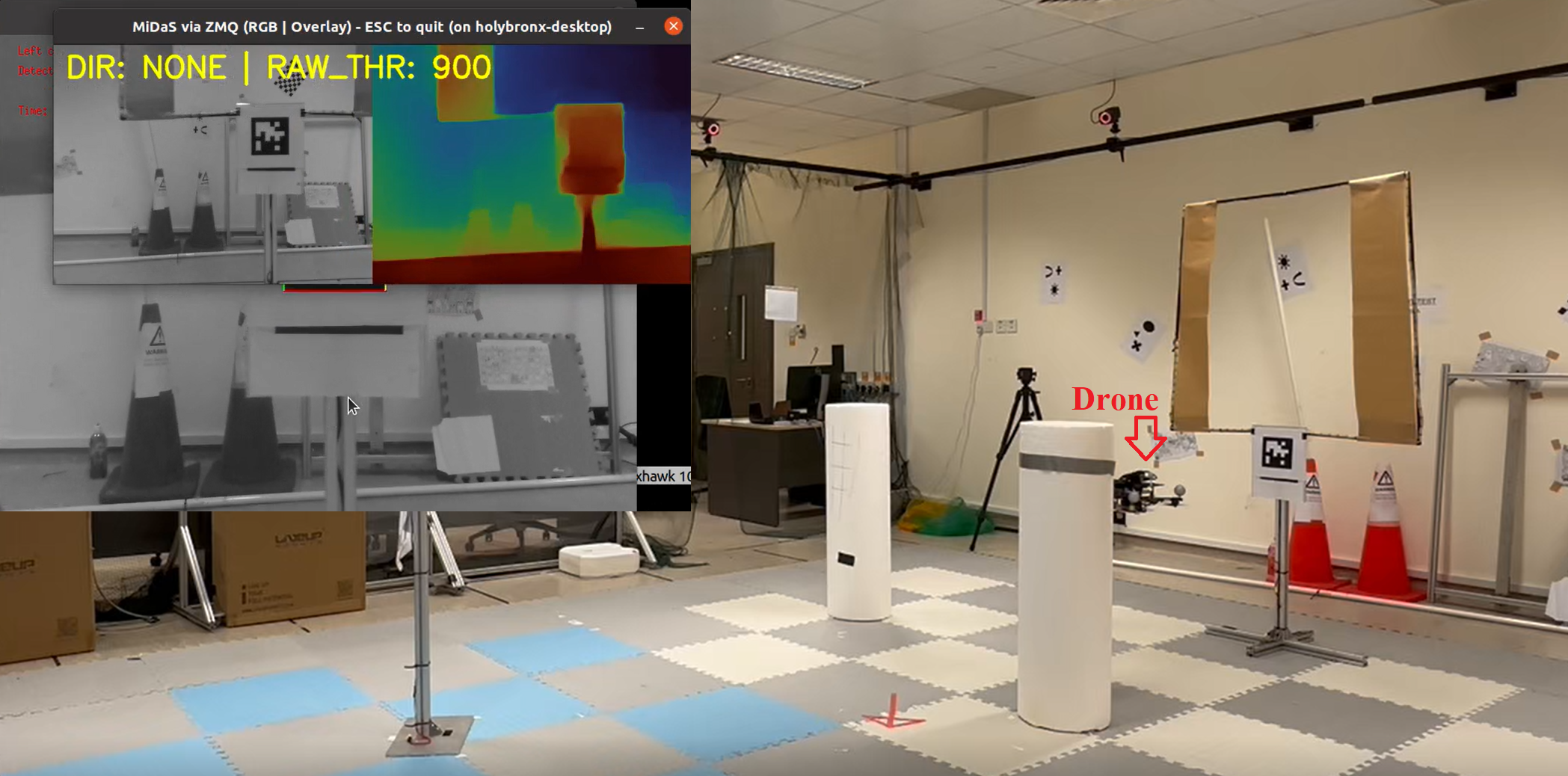}
        \caption{UAV avoids the obstacle}
        \label{fig:colission_avoidance}
    \end{subfigure}

    \caption{Navigation and obstacle avoidance process: (a) UAV tracks the first AprilTag; (b) UAV crosses the gate; (c) UAV tracks the second AprilTag; (d) UAV avoids the obstacle.}
    \label{fig:ibvs_midas}
\end{figure*}

\item \underline{Crossing the Gate}: The drone first ascends vertically until it reaches the center of the gate. It then flies forward to complete the crossing.
Afterward, it descends to the original flight altitude, preparing to track the second AprilTag.

\item \underline{MiDaS-based Obstacle Avoidance}: During the tracking of the second AprilTag, the UAV encountered obstacles along its flight path. To ensure safe navigation, the MiDaS module processed RGB images transmitted from the IBVS via \texttt{zmq\_tcp} and generated relative depth maps in real time. Based on these predictions, regions with relative depth values greater than a predefined threshold were identified (since larger MiDaS depth values indicate shorter distances, such regions correspond to nearby obstacles). The spatial distribution of these regions was then analyzed to determine whether the obstacles were located on the left or right side of the UAV’s forward direction. Accordingly, the IBVS controller executed avoidance actions, allowing the UAV to bypass the obstacles while maintaining continuous tracking of the AprilTags.

\item \underline{Tracking the $2^{\mathrm{nd}}$ AprilTag}: When no obstacles were detected in front of the UAV by the MiDaS module, the AprilTag served as the only navigation reference. In this case, the UAV continued to track the second AprilTag using the IBVS framework, with the image error gradually decreasing as the vehicle aligned with the target. This ensured stable navigation toward the desired gate.

\end{itemize}

\begin{figure*}[t]
    \centering
    \begin{subfigure}[b]{0.49\linewidth}
        \centering
        \includegraphics[width=\linewidth]{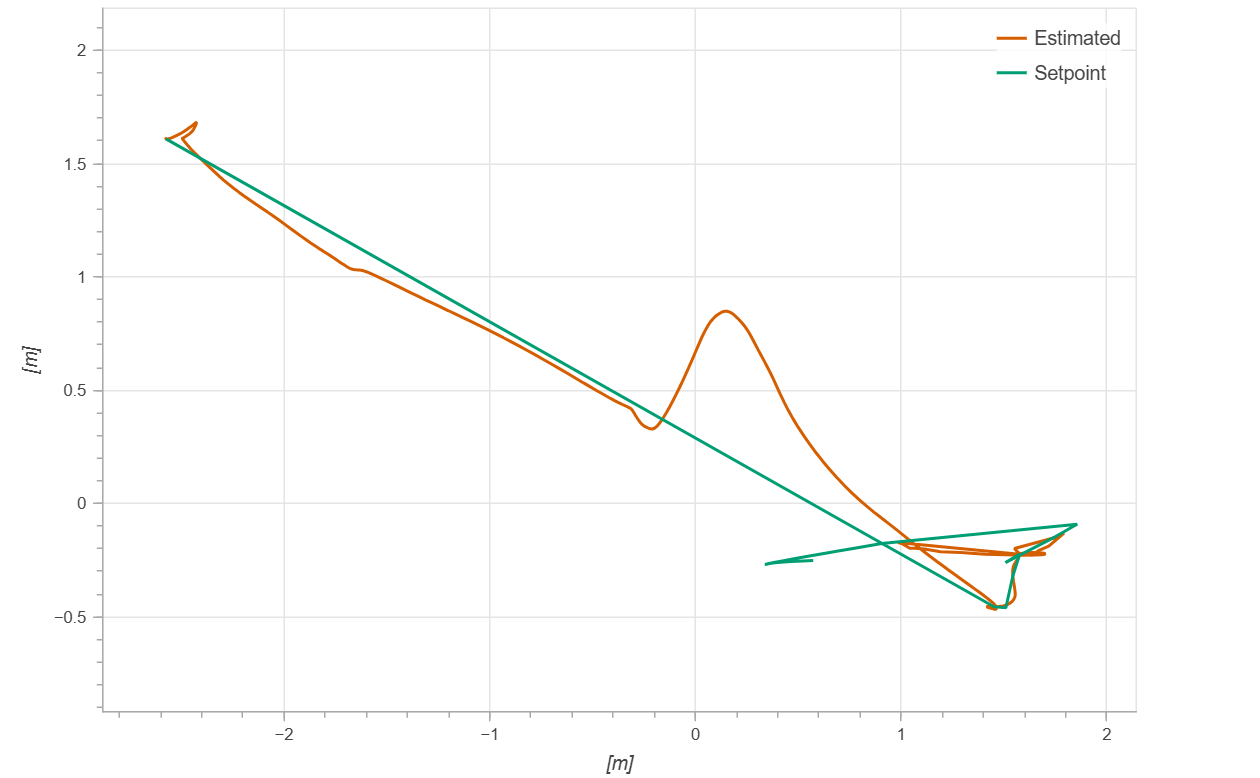}
        \caption{Trajectory with IBVS+MiDaS}
        \label{fig:position_midas}
    \end{subfigure}
    \begin{subfigure}[b]{0.49\linewidth}
        \centering
        \includegraphics[width=\linewidth]{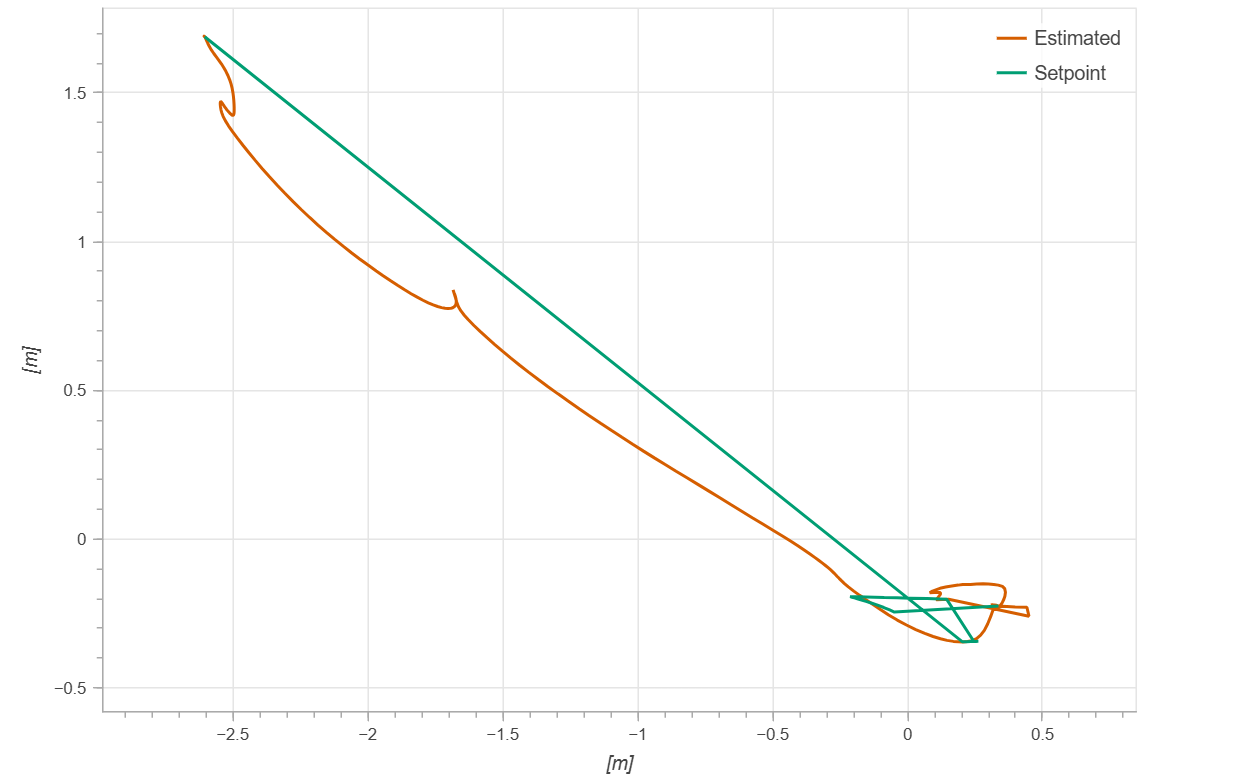}
        \caption{Trajectory with IBVS only}
        \label{fig:position_only}
    \end{subfigure}

    \begin{subfigure}[b]{0.49\linewidth}
        \centering
        \includegraphics[width=\linewidth]{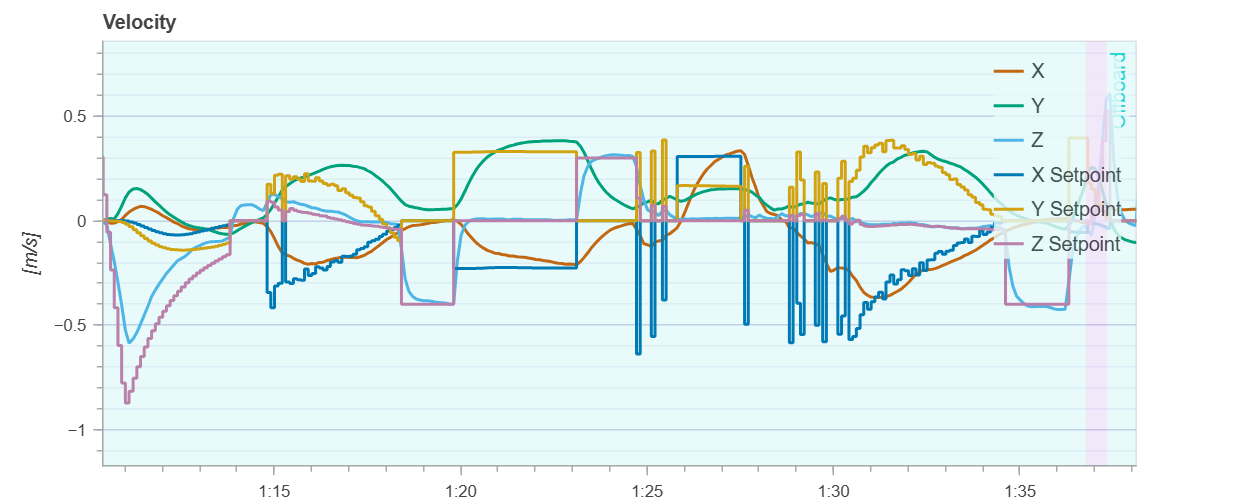}
        \caption{Linear velocity tracking with IBVS+MiDaS}
        \label{fig:velocity_midas}
    \end{subfigure}
    \begin{subfigure}[b]{0.49\linewidth}
        \centering
        \includegraphics[width=\linewidth]{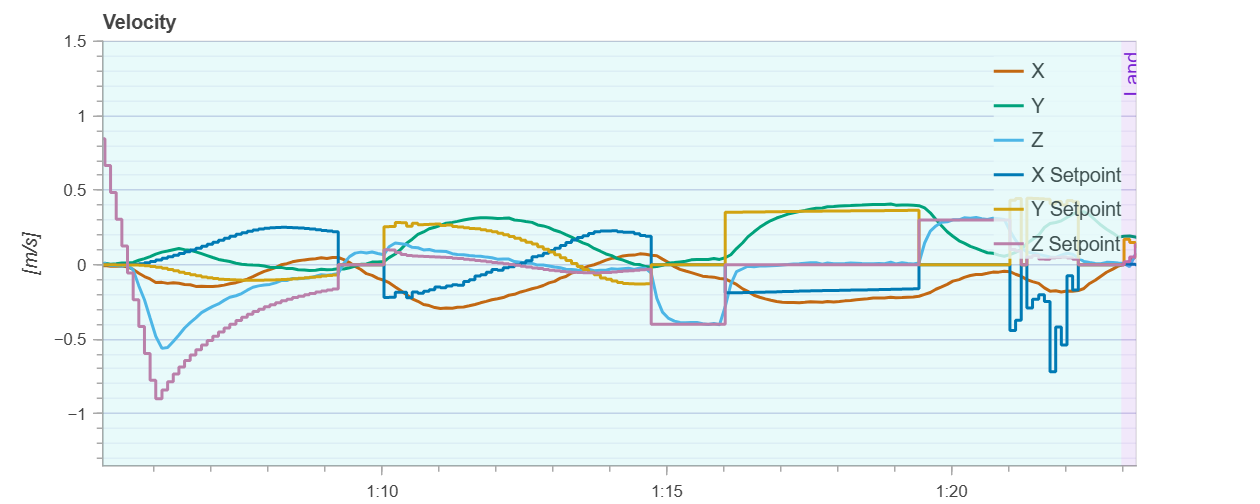}
        \caption{Linear velocity tracking with IBVS only}
        \label{fig:velocity_only}
    \end{subfigure}

    \begin{subfigure}[b]{0.49\linewidth}
        \centering
        \includegraphics[width=\linewidth]{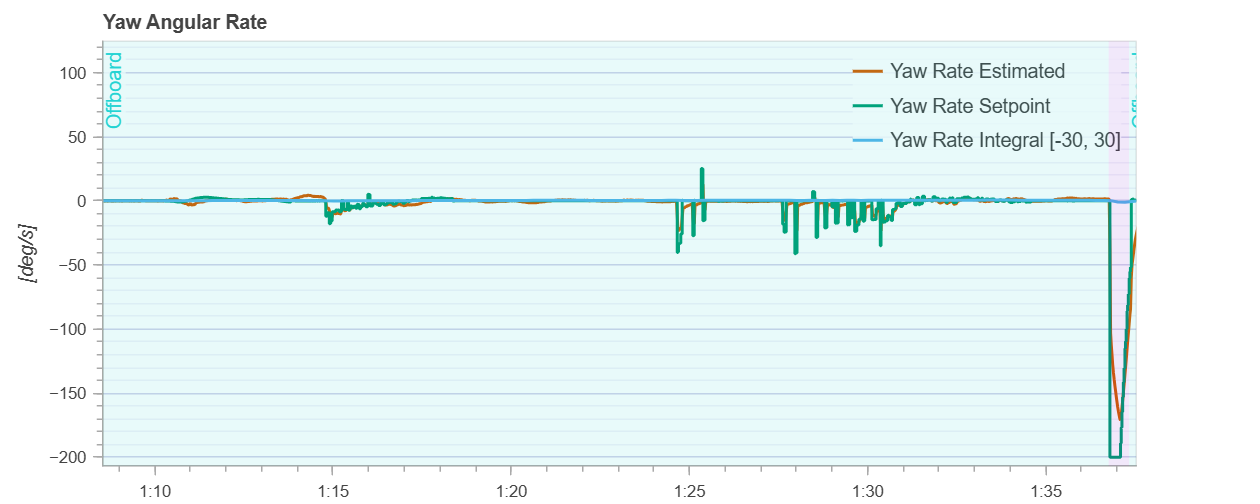}
        \caption{Yaw rate  with IBVS+MiDaS}
        \label{fig:yaw_midas}
    \end{subfigure}
    \begin{subfigure}[b]{0.49\linewidth}
        \centering
        \includegraphics[width=\linewidth]{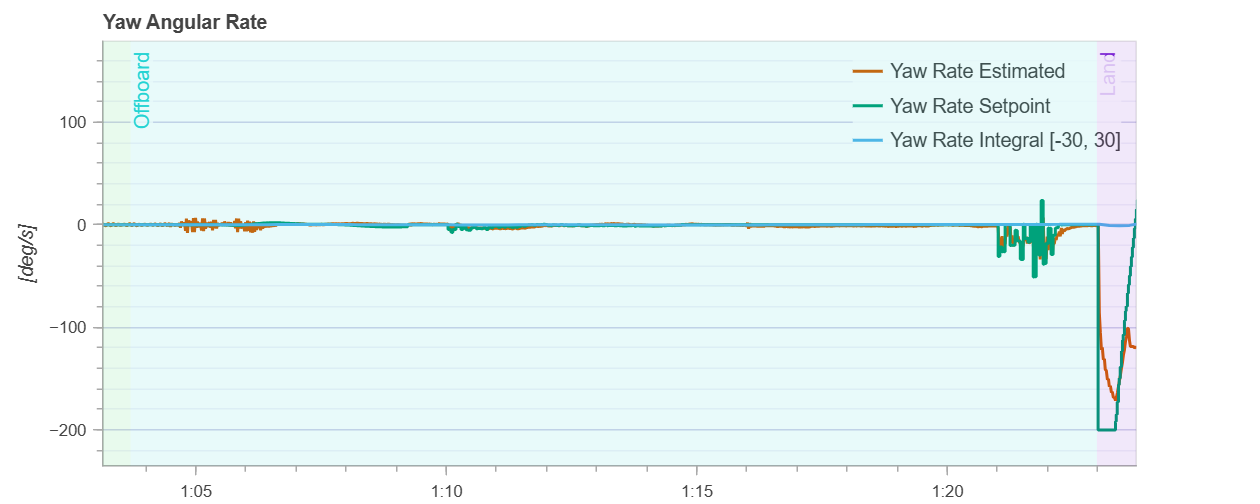}
        \caption{Yaw rate with IBVS only}
        \label{fig:yaw_only}
    \end{subfigure}

    \caption{UAV tracking performance: (a)(b) position trajectories with and without MiDaS; (c)(d) linear velocity tracking in X, Y, Z axes with and without MiDaS; (e)(f) yaw rates during two experiments with and without MiDaS}

    \label{fig:tracking_compare}
\end{figure*}

As illustrated in Fig.~\ref{fig:tracking_compare}, the UAV completed the full mission of target tracking and obstacle avoidance. The position trajectory recorded in the VICON system is shown in Fig.~\ref{fig:tracking_compare}a. When encountering obstacles, necessary deviations were performed, after which the UAV returned to the reference trajectory and reached the target. 
The linear velocity profiles (Fig.~\ref{fig:tracking_compare}c) further confirm that the UAV was able to follow the IBVS reference commands in the X, Y, and Z axes. Although temporary fluctuations occurred during the avoidance maneuver, the velocities remained within controllable bounds. 
In addition, the yaw rates recorded during two experiments (Fig.~\ref{fig:tracking_compare}e--f) indicate that the UAV maintained stable heading control while performing gate crossing and obstacle avoidance, with only minor deviations from the reference yaw rate setpoints. These results demonstrate that the combination of IBVS with MiDaS enables reliable target tracking, stable rotation regulation, and effective obstacle avoidance in cluttered environments.

\subsection{Comparison of experimental results with existing literature}

\subsubsection{IBVS without use of  collision avoidance}

\begin{figure}[thpb]
    \centering
    \includegraphics[width=0.85\linewidth]{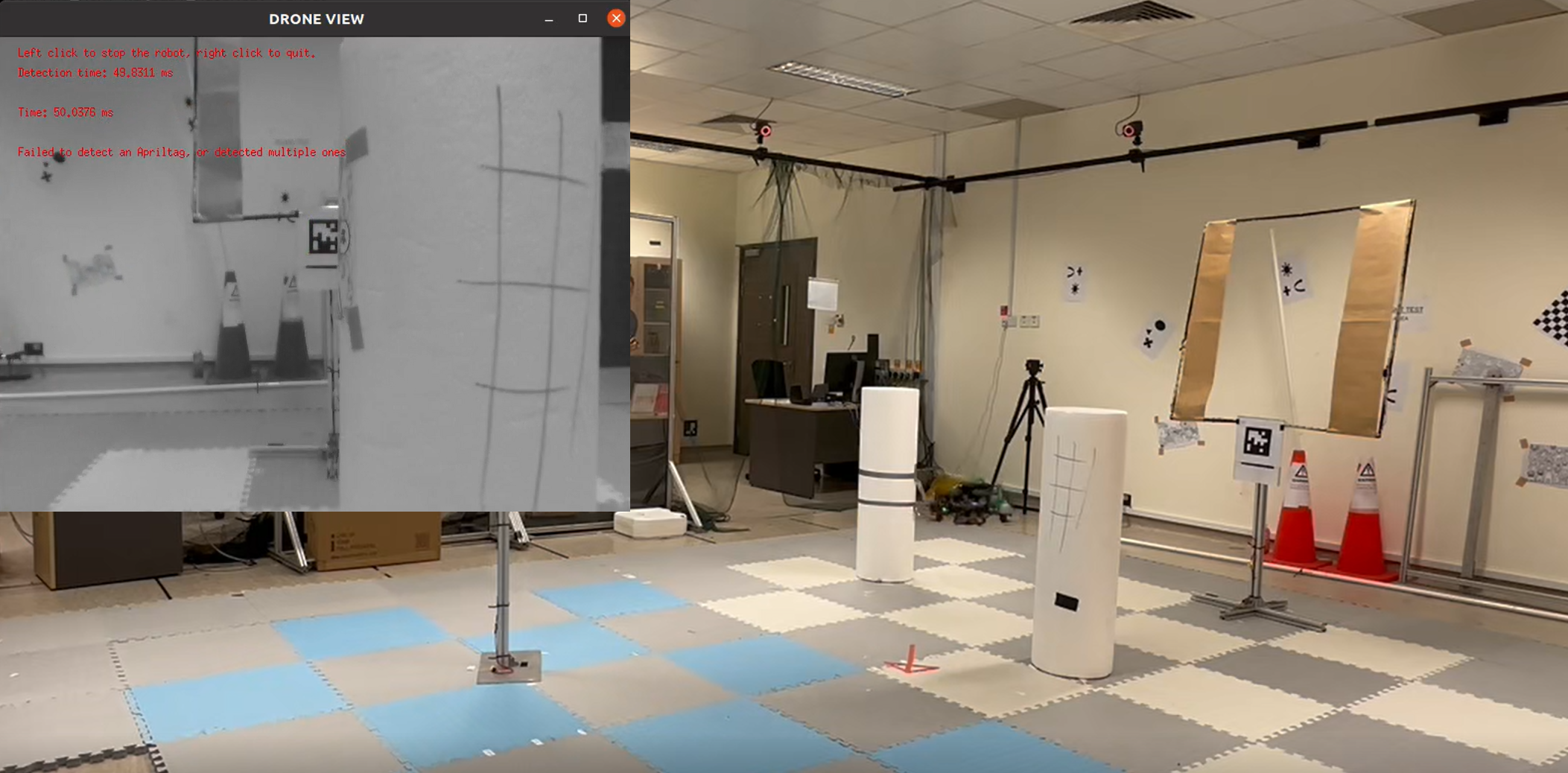}
    \caption{UAV will collide with the obstacle}
    \label{fig:obstacle}
\end{figure}

In the absence of MiDaS-based depth estimation, the UAV relied solely on the IBVS framework for navigation. After successfully detecting and tracking the first AprilTag, the UAV was able to pass through the initial gate. However, during the subsequent stage of tracking the second AprilTag, the UAV encountered cylindrical obstacles along its flight path. Since no relative depth information was available, the obstacles were not perceived in advance. As a result, the UAV either lost sight of the AprilTag due to visual occlusion or collided with the obstacle, leading to mission failure. These results highlight the necessity of integrating MiDaS depth estimation into IBVS to ensure safe and reliable navigation in cluttered environments.

Without the support of MiDaS depth estimation, the UAV relied solely on IBVS for navigation. As shown in Fig.~\ref{fig:position_only}, the UAV was able to detect and track the first AprilTag and complete the gate-crossing maneuver. However, when proceeding toward the second AprilTag, the UAV failed to perceive the obstacles along its flight path. Consequently, it continued moving forward as computed (Fig.~\ref{fig:tracking_compare}b), but due to the occlusion caused by the obstacles, the UAV eventually lost sight of the target or faced potential collision. The velocity profiles (Fig.~\ref{fig:tracking_compare}d) further illustrate this phenomenon: the fluctuations in the $X$, $Y$, and $Z$ directions reflected the UAV’s attempts to follow the new AprilTag, but as the UAV approached the obstacle too closely, manual intervention was required to land the vehicle and avoid collision safely.

In contrast, when MiDaS was integrated into the IBVS framework (Fig.~\ref{fig:tracking_compare}), the UAV was able to sense nearby obstacles through relative depth maps and execute appropriate avoidance maneuvers. Although short-term fluctuations in velocity were observed during avoidance, they remained within controllable bounds, and the UAV successfully returned to the target without losing it. This comparison clearly demonstrates that MiDaS-assisted IBVS significantly improves the robustness and safety of UAV navigation in cluttered environments.

\subsubsection{Several obstacle detection methods with a simulated depth camera}

Existing literature \cite{ hz_jz_xy_yl_lc_kz_sl_xh_2023,ly_xw_yz_zl_ls_2024,dl_hl_hjk_2011} uses the exact depth information for path planning or collision avoidance, and their results are based on simulation tests. In this part, we will show that it is difficult to use the depth information to get the exact 3D positions for collision avoidance. 

\begin{figure}[t]
    \centering
    \begin{subfigure}[b]{0.48\linewidth}
        \centering
        \includegraphics[width=\linewidth,height=2.2cm]{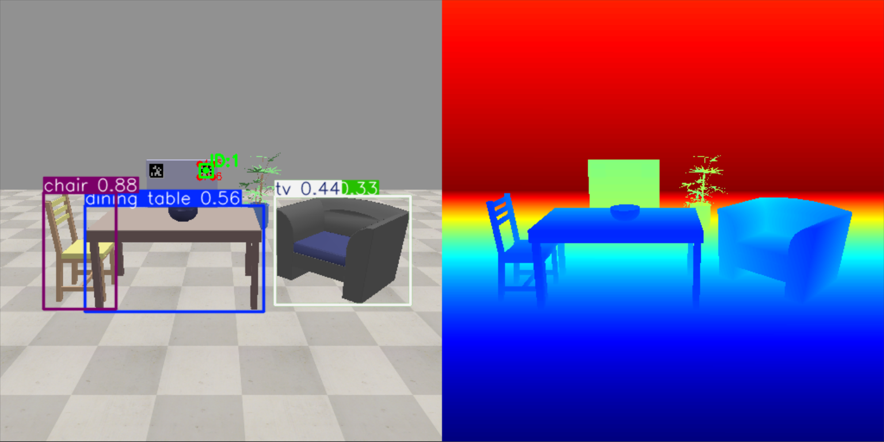}
        \caption{Simulated depth camera}
        \label{fig:sim1}
    \end{subfigure}
    \hfill
    \begin{subfigure}[b]{0.48\linewidth}
        \centering
        \includegraphics[width=\linewidth,height=2.2cm]{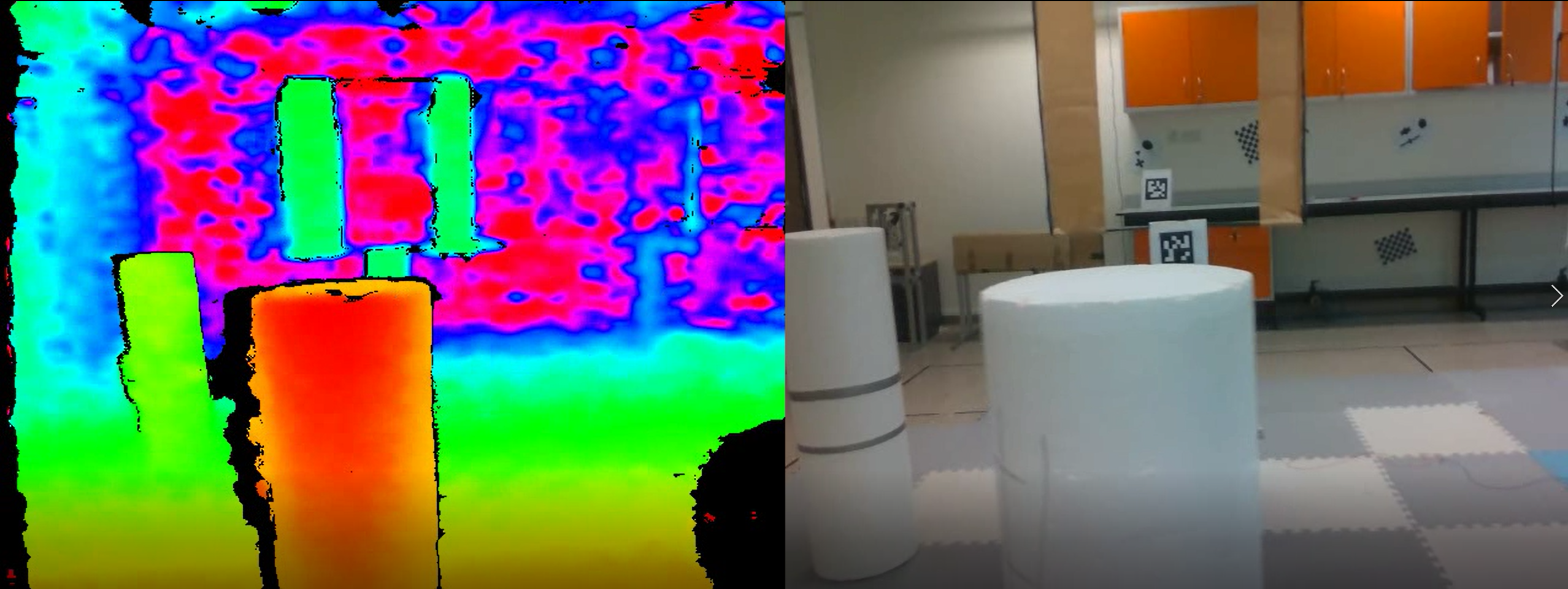}
        \caption{D435i depth camera}
        \label{fig:d435i}
    \end{subfigure}
    \caption{Comparison of depth perception sources: (a) simulated depth camera; (b) Intel RealSense D435i depth camera stream.}
    \label{fig:depth_compare}
\end{figure}

To further investigate the impact of the depth camera on obstacle detection, we compared a simulated depth camera in CoppeliaSim (Figure~\ref{fig:sim1}) with the Intel RealSense D435i depth camera (Figure~\ref{fig:d435i}) equipped on our drone.
The simulated depth sensor provides depth maps with very high resolution, producing smooth and noise-free measurements across the entire pre-designed sensing range. Consequently, obstacles are clearly visible with sharp boundaries, ensuring that obstacle detection can be reliably achieved throughout the measurement range.

In contrast, the D435i depth camera suffers from several practical limitations.
Firstly, its effective working range is restricted to approximately \textbf{3\,m -- 4\,m for optimal accuracy}, with usable but less reliable measurements up to around 10\,m. This limitation causes constraints on UAV operations in environments where obstacles may appear beyond the optimal sensing distance. 
Secondly, due to the stereo-based sensing principle, the D435i produces lower resolution depth images compared to the simulated sensor, and depth estimation quality degrades significantly near image boundaries. In these regions, stereo correspondence often fails, resulting in missing or invalid measurements.  

These shortcomings motivate us to adopt \textbf{learning-based monocular depth estimation} using MiDaS. Unlike the stereo camera, MiDaS can infer relative depth from a single RGB image, providing smoother and more complete depth maps even in regions where stereo cameras typically fail. Moreover, MiDaS isn't limited by the optimal working range, allowing obstacle detection to generalize across a wider variety of environments. This explains why our IBVS framework combines MiDaS-based depth prediction as a replacement for the hardware depth camera in UAV collision avoidance tasks.

\begin{figure}[thpb]
    \centering
    \includegraphics[width=0.85\linewidth]{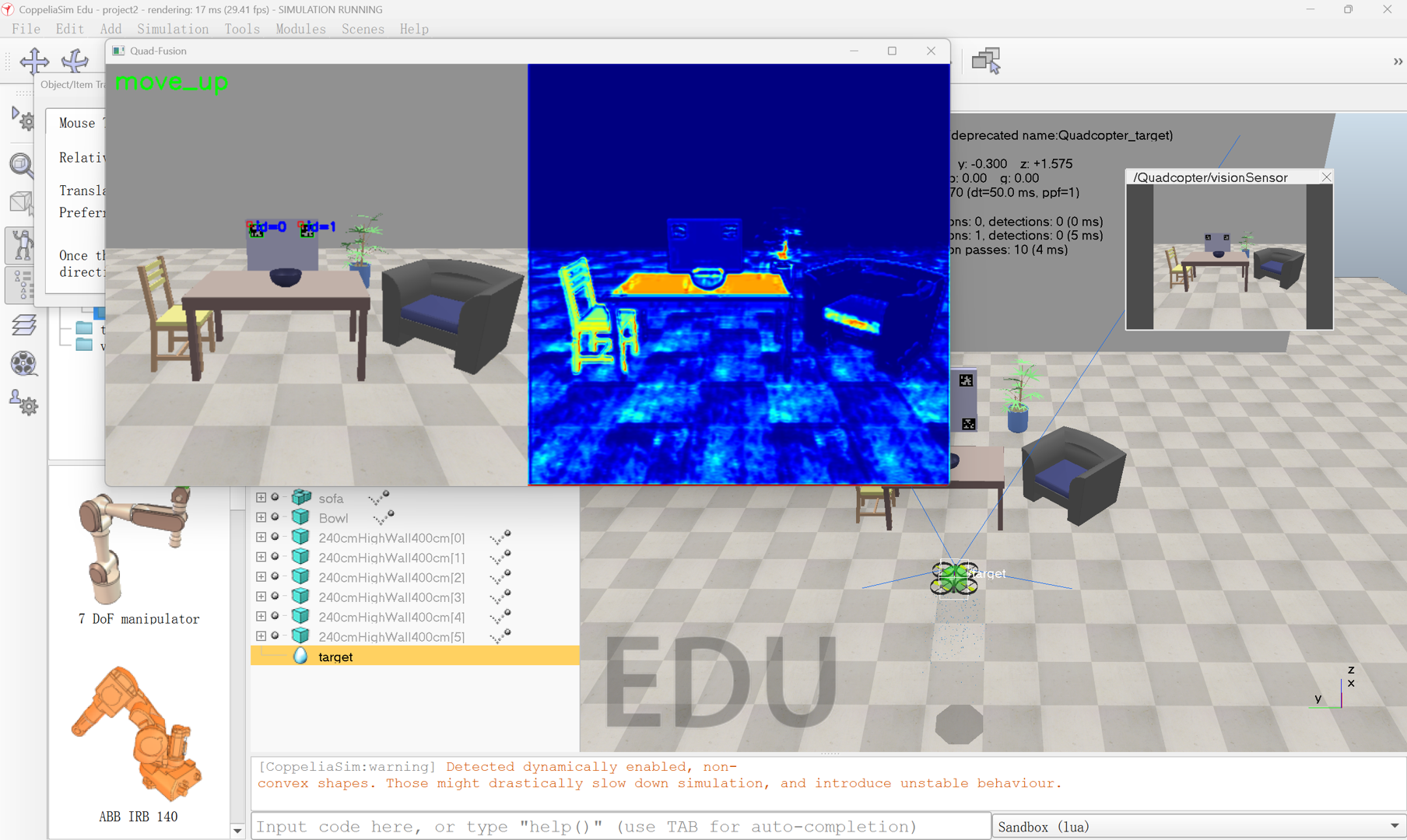}
    \caption{The fusion model with small parameters gives obstacle prompts}
    \label{fig:sim2}
\end{figure}

As illustrated in Fig.~\ref{fig:sim2}, we first designed and tested a \emph{small-parameter fusion model} by combining YOLOv8, optical flow, and GLCM (Gray-Level Co-occurrence Matrix) texture descriptors in the CoppeliaSim environment. The rationale behind this design was to maintain a lightweight computation cost while still capturing complementary cues from appearance, motion, and texture. 

The simulation results demonstrate that this fusion model can provide coarse obstacle prompts. 
In the CoppeliaSim environment, the simulated depth camera is noise-free, the illumination is stable, the optical flow field remains smooth, and the GLCM descriptors clearly reflect the object boundaries. 
However, in real-world scenarios, in addition to the aforementioned limitations of depth cameras, varying illumination conditions, motion blur during flight, flickering, and high dynamic range all prevent this model from being directly applicable to practical obstacle detection.

\begin{figure}[t]
    \centering
    \begin{subfigure}[b]{0.48\linewidth}
        \centering
        \includegraphics[width=\linewidth]{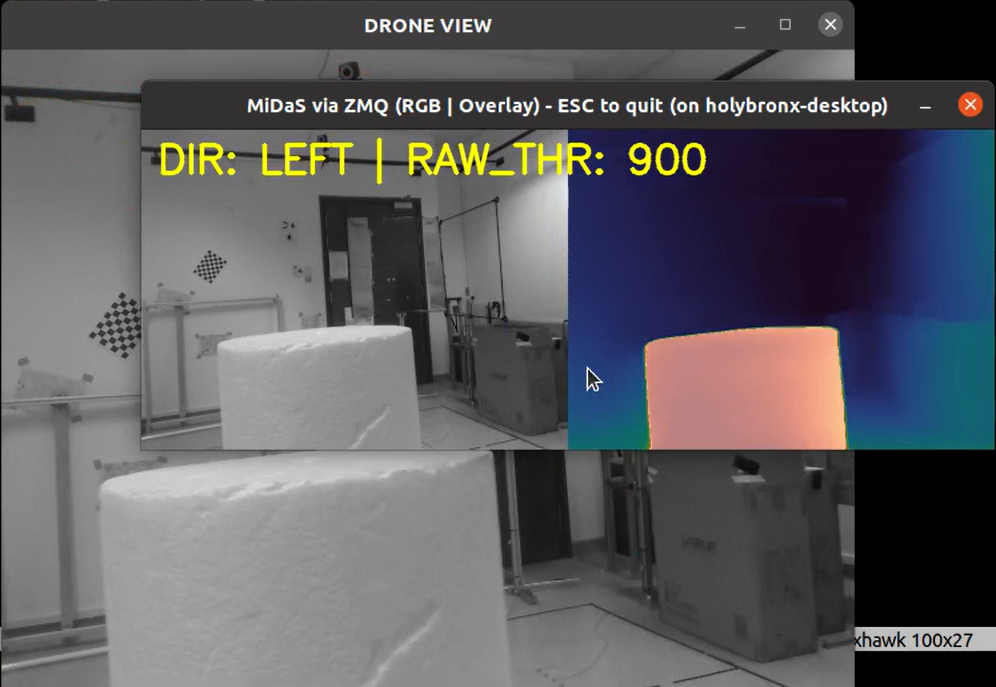}
        \caption{MiDaS Visualization (Left obstacle)}
        \label{fig:left_vis}
    \end{subfigure}
    \hfill
    \begin{subfigure}[b]{0.48\linewidth}
        \centering
        \includegraphics[width=\linewidth]{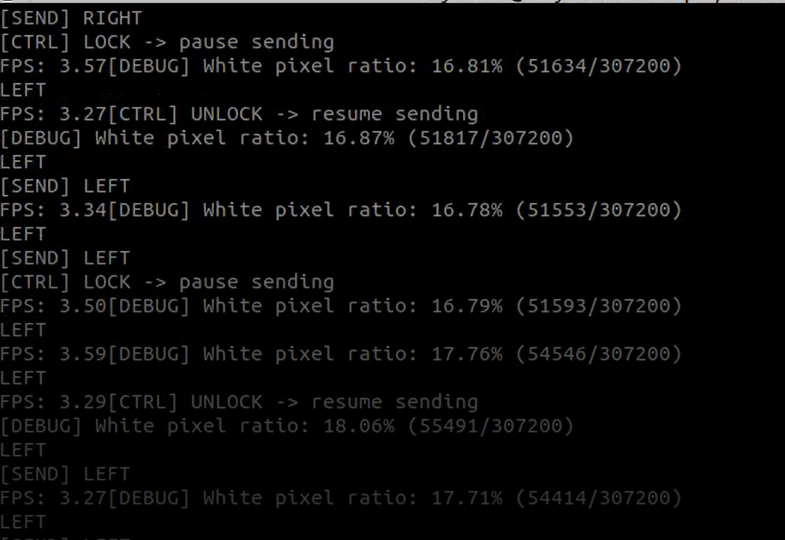}
        \caption{Avoidance action (Left obstacle)}
        \label{fig:left_com}
    \end{subfigure}

    \begin{subfigure}[b]{0.48\linewidth}
        \centering
        \includegraphics[width=\linewidth]{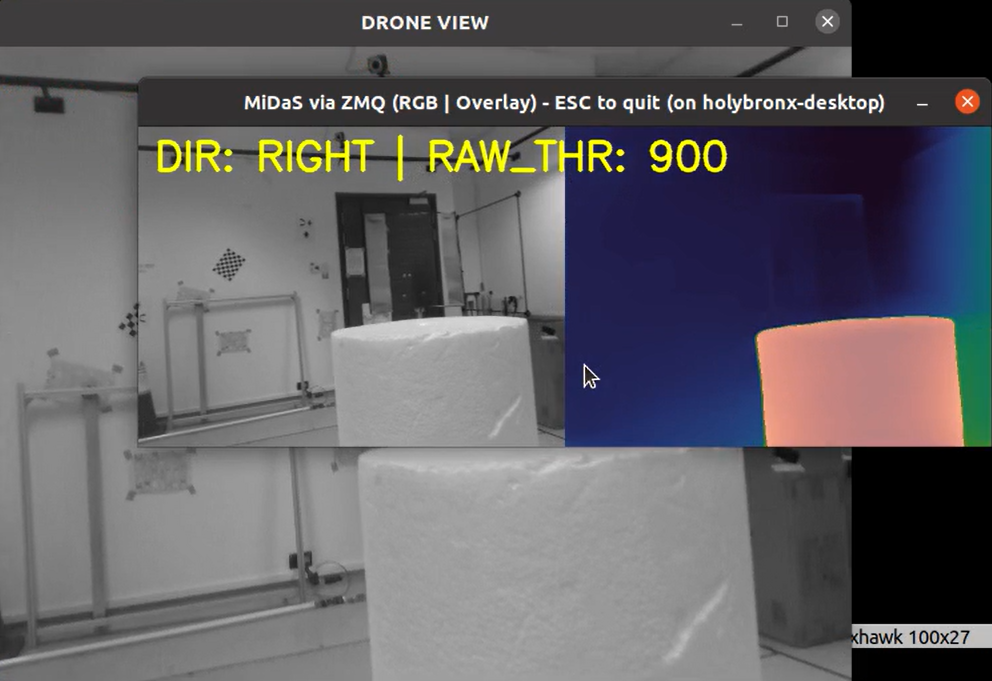}
        \caption{MiDaS Visualization (Right obstacle)}
        \label{fig:right_vis}
    \end{subfigure}
    \hfill
    \begin{subfigure}[b]{0.48\linewidth}
        \centering
        \includegraphics[width=\linewidth]{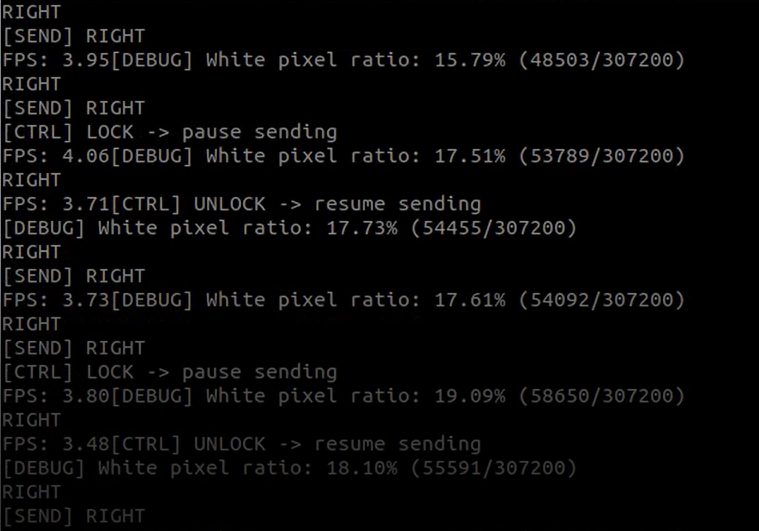}
        \caption{Avoidance action (Right obstacle)}
        \label{fig:right_com}
    \end{subfigure}

    \caption{MiDaS-based obstacle avoidance motions: (a)(b) left-side obstacle case; (c)(d) right-side obstacle case.}
    \label{fig:midas_lr}
\end{figure}

In contrast, our final choice---\textbf{MiDaS-based monocular depth}---directly regresses a dense \emph{relative} depth map from a single RGB frame. We adopt the \emph{OpenVINO} version of \texttt{midas\_v21\_small\_256}, whose parameter count is about \textbf{16.6M}, enabling real-time inference on edge devices while preserving robust cross-domain generalization. As illustrated in Fig.~\ref{fig:midas_lr} (ab) and Fig.~\ref{fig:midas_lr} (cd), MiDaS provides smooth, complete relative depth even in low-texture areas and around image borders; we then normalize and threshold the inverse depth to mark \textbf{near-field obstacles}, and, via our designed rule-based policy, issue LEFT/RIGHT avoidance commands to the IBVS controller (with LOCK/UNLOCK handshake) to execute safe maneuvers.

We compare the navigation methods considered in this paper.
In summary, we show Table \ref{table_example2}, giving some information regarding each method: camera used, can/cannot do navigation and collision avoidance. It can be seen that the proposed method can achieve navigation with collision avoidance using a low-cost RGB camera in a real flight test environment. 

\begin{table*}[h]
\caption{A comparison of several methods: Y represents 'Yes'; $\times$ represents 'Cannot Do'. }
\label{table_example2}
\begin{center}
\begin{tabular}{|c||c||c||c|}
\hline
Methods & Cameras & Navigation & Collision Avoidance\\
\hline
Proposed method & RGB & Y & Y \\
\hline
IBVS~\cite{cdw_fpv_ns_gched_2022} & stereo cameras & Y & $\times$ \\
\hline
Stereo camera based navigation~\cite{cq_msjm_jc_hhtl_2024} & stereo cameras & Y & $\times$ \\
\hline
Stereo camera with path planning / predictive control~\cite{hz_jz_xy_yl_lc_kz_sl_xh_2023,ly_xw_yz_zl_ls_2024,dl_hl_hjk_2011} & stereo cameras & Y & ? \\
\hline
\end{tabular}
\end{center}
\end{table*}

Here, ``$\times$'' indicates that the method cannot perform collision avoidance, 
as these approaches rely solely on stereo depth or visual servoing for navigation 
without incorporating obstacle detection or avoidance strategies. The mark ``?'' denotes that although these stereo camera-based methods introduce path planning or predictive control verified by simulation, their capability for reliable collision avoidance has not been conclusively validated in the real world. Most reported results are limited to simulation or constrained environments, leaving their practical effectiveness in real-world scenarios uncertain.

\section{CONCLUSIONS}

This paper presented an IBVS-based navigation and collision avoidance framework that operates solely with a low-cost RGB camera. By combining AprilTag-based visual servoing with MiDaS monocular depth estimation, the system enables UAVs to track sequential visual targets and perform safe obstacle avoidance without relying on GPS, LiDAR, or stereo vision.

A key feature of our work is that \textbf{all computations are executed on-board the Jetson platform}, including the MiDaS depth inference. Unlike many existing approaches that offload heavy perception tasks to external high-performance workstations, our system remains fully self-contained, thereby eliminating communication latency and external dependencies. This makes our framework both more practical and deployable in real-world scenarios.

Experimental results demonstrate that the UAV successfully tracks multiple AprilTags, crosses designated gates, and avoids obstacles in real time. The comparison with baseline IBVS-only navigation shows that MiDaS-assisted perception significantly improves robustness and safety. Furthermore, the ability to run the entire pipeline onboard highlights the feasibility of deploying AI-driven perception and control on compact aerial platforms.

In future work, we will extend the framework to more complex environments with dynamic obstacles and evaluate the scalability of the onboard perception-control loop. We also aim to explore lightweight network compression and sensor fusion strategies to further enhance the efficiency and reliability of the system.

\addtolength{\textheight}{-12cm}   




\section*{ACKNOWLEDGMENT}

The authors would like to thank Derek Neo for the drone design shown in Fig.~\ref{fig:drone} and his support during flight tests.


\end{document}